\documentclass{article}

\usepackage{multirow}
\usepackage{subcaption}
\usepackage{graphicx}
\usepackage[nonatbib,final]{neurips_ts4h_2022}
\usepackage[numbers]{natbib}

\usepackage[utf8]{inputenc} \usepackage[T1]{fontenc}    \usepackage{hyperref}       \usepackage{url}            \usepackage{booktabs}       \usepackage{amsfonts}       \usepackage{nicefrac}       \usepackage{microtype}      \usepackage{xcolor}         \usepackage{wrapfig}
\usepackage{subcaption}
\newcommand{\m}{\mathbf}

\title{On the Importance of Clinical Notes in Multi-modal Learning for EHR Data}

\author{Severin Husmann
  \And Hugo Yèche
  \And Gunnar Rätsch \thanks{Co-supervised}
  \And Rita Kuznetsova \footnotemark[1] \\
  \\
  \\
  \texttt{\{shusmann,hyeche,raetsch,mkuznetsova\}@ethz.ch} \\
  Department of Computer Science, ETH Zürich \\
}

\begin{document}

\maketitle

\begin{abstract}
Understanding deep learning model behavior is critical to accepting machine learning-based decision support systems in the medical community. Previous research has shown that jointly using clinical notes with electronic health record (EHR) data improved predictive performance for patient monitoring in the intensive care unit (ICU). In this work, we explore the underlying reasons for these improvements. While relying on a basic attention-based model to allow for interpretability, we first confirm that performance significantly improves over state-of-the-art EHR data models when combining EHR data and clinical notes. We then provide an analysis showing improvements arise almost exclusively from a subset of notes containing broader context on patient state rather than clinician notes. We believe such findings highlight deep learning models for EHR data to be more limited by partially-descriptive data than by modeling choice, motivating a more data-centric approach in the field.
\end{abstract}

\section{Introduction} 
Recently, researchers have explored using deep neural networks for clinical decision support, particularly for Intensive Care Unit (ICU) prediction tasks. Most existing literature focused on using either available electronic health records~\cite{futoma2017learning,tomavsev2019clinically,schwab2020real, horn20, li2020time, hyland2020early,yeche2021} (EHR) or clinical notes~\cite{grnarova2016neural, huang19, feng2020explainable, vanaken21} individually. However, more recent research shows that combining both data types in a cross-modal fashion improves performance~\cite{khadanga19, chen20} on a popular suite of MIMIC-III~\cite{mimiciii} benchmark tasks~\cite{harutyunyan19}. Ultimately, the goal of such improvement in the field can help better plan the resource usage in ICU~\cite{khadanga19, bates14}. Though, as for any real-world application, acceptance by clinicians will require an understanding of the behavior of these new cross-modal models. Other studies provided analyses for uni-modal models using EHR data~\cite{horn20} or clinical notes~\cite{grnarova2016neural}. To the best of our knowledge, similar work investigating the cross-modal interactions when combining both modalities to monitor patients in ICU does not exist.

We aim to fill that gap by exploring the underlying reasons for performance improvements obtained from jointly using EHR data and clinical notes. Like \citet{horn20}, which analyzed the performances of EHR-only models, we rely on a transformer-based~\citep{vaswani17} model, as attention weights provide a direct explanation of the deep learning model behavior. Due to the multi-modal nature of the data, we adopt a multi-modal extension~\citep{tsai19} to this original transformer architecture. In this work, first, we confirm that the most basic cross-modal transformer model combining EHR data and clinical notes performs significantly better than state-of-the-art uni-modal approaches on all classification tasks from the MIMIC-III benchmark~\citep{harutyunyan19}. Second, by performing a two-step analysis, we show that the performance gain compared to EHR-only models arises exclusively from a subset of nurses' notes and radiology reports. These notes provide additional context on patients' states. Moreover, we also show that using clinicians' notes, which summarize EHR contents rather than containing additional information, does not provide any performance gain. Thereby, we refute the hypothesis that current architectures poorly represent EHR information. We believe our findings motivate more data-centric approaches in the field, focusing on leveraging more descriptive data to model patients' states in the ICU.

\section{Related work}\label{papers}

\paragraph{Deep Learning for ICU patient monitoring.}
As previously mentioned, we can categorize existing research using deep learning to monitor patients' evolution in the ICU according to the modality used. First, the most common line of work relies on the EHR, viewed as a multivariate time series. A large group of related literature tackles this problem with various architectures in both supervised~\cite{futoma2017learning,tomavsev2019clinically,schwab2020real, horn20, li2020time, hyland2020early,yeche2021} and unsupervised fashion~\cite{lyu2018improving, yeche2021neighborhood}. Other researchers focus exclusively on clinical notes~\cite{grnarova2016neural, huang19,feng2020explainable, vanaken21}. At last, motivating our work, more recent research explored fusing both modalities~\cite{khadanga19, chen20, jin2018improving, yang2021multimodal}, yielding significant improvement over previous uni-modal approaches. Unfortunately, in this last body of work, the authors work with independent representations for each modality making it difficult to retrieve cross-modal interaction, thus limiting interpretability.

\paragraph{Interpretability in deep learning.}
Widely used methods in deep learning interpretability rely on saliency-map-based methods like LIME~\cite{ribeiro16}, GRAD-CAM~\cite{woo2018cbam}, and SHAP~\cite{lundberg17}. Though, these methods are limited~\cite{adebayo2018sanity} and are not necessarily compatible with sequence data. Thus, we base our analysis on attention-based methods, a more common approach for sequence modeling. Various papers have investigated the use of attention weights as a potential post hoc interpretability mechanism~\cite{clark19, vig19, kobayashi20, rogers20}. In addition, other literature further developed methods beyond attention weights, such as attention flow and rollout~\cite{abnar20}. In the case of attention flow, \cite{ethayarajh21} proved that attention flows are layer-wise Shapley values. Attention rollout is often employed in computer vision to provide attention heatmaps over the input~\cite{kolesnikov21, nagrani21}. In the context of ICU data, similarly to us, \citet{horn20} also based their uni-modal model behavior analysis on attention weights.

\section{Method}\label{method}
In this section, we introduce our interpretable model to handle both clinical notes and EHR data. We also describe hyperparameter choice and data pre-processing to ensure a better analysis of clinical notes' importance.

\paragraph{Architecture.}
 \begin{wrapfigure}[15]{i}{0.3\textwidth}
 \vspace{-1.5em}
  \begin{center}
    \includegraphics[height=4.2cm,scale=1]{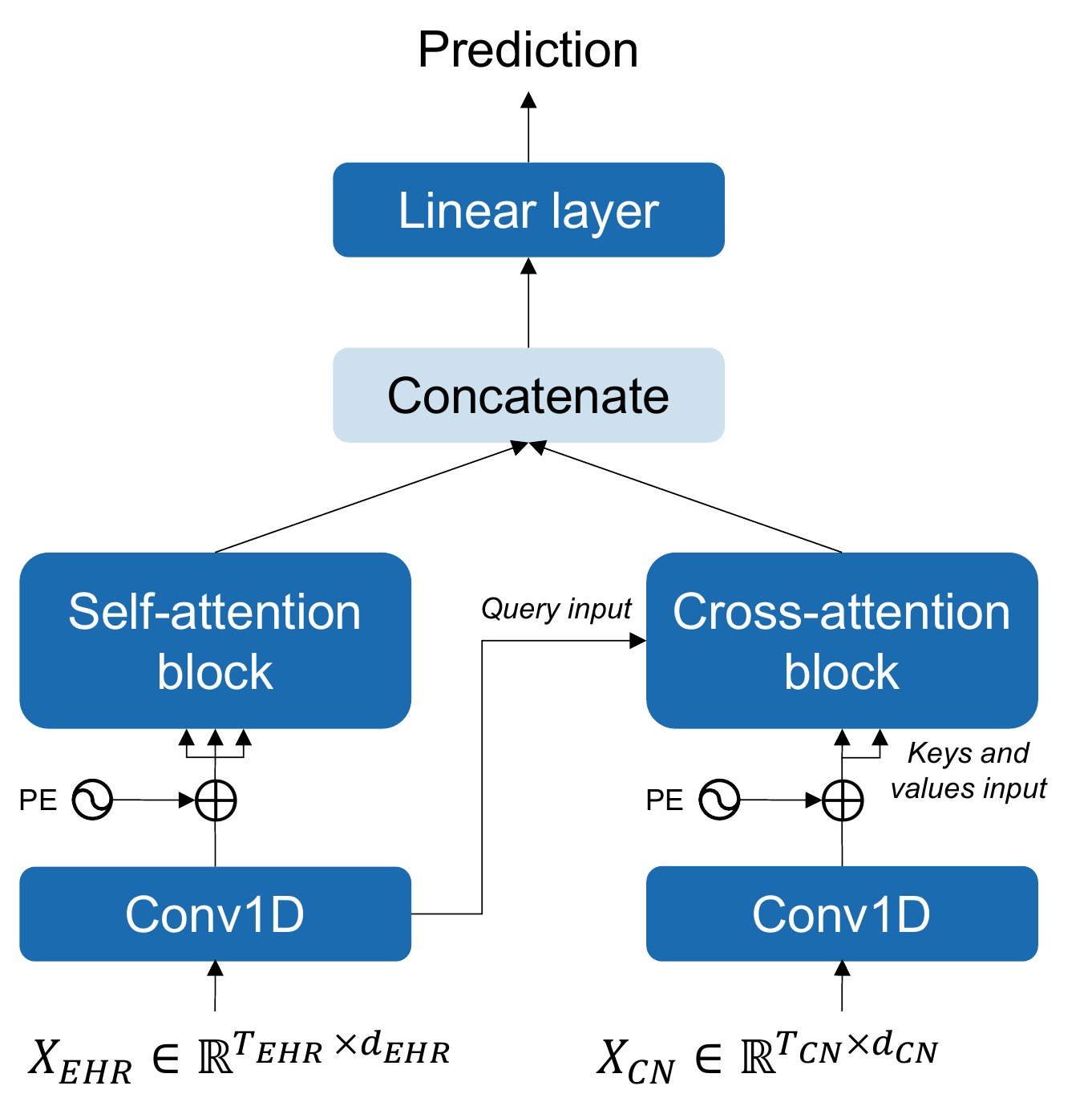}
  \end{center}
  \caption{Proposed cross-modal architecture}
  \label{fig:architecture}
\end{wrapfigure}
We consider two modalities $\m{X}_{EHR} \in \mathbb{R}^{T_{EHR} \times d_{EHR}}$ and $\m{X}_{CN} \in \mathbb{R}^{T_{CN} \times d_{CN}}$, where $T_{EHR}$ and $T_{CN}$ denote the length of the EHR data and clinical notes during the given patient's stay, and $d_{EHR}$ and $d_{CN}$ the input feature dimensionality. To obtain $\m{X}_{CN}$ we use the pre-trained Clinical BERT model~\cite{alsentzer19} as a text feature extractor, $d_{CN} = 769$. Note that $d_{CN}$ is greater than the original Clinical BERT output dimension by one. This is because clinical notes are not regularly spaced, which cannot be accounted for by positional encoding. Thus, we add the entry time of notes as the last feature. The dimensionality of EHR data is $d_{EHR} = 42$ after one-hot encoding of categorical variables. Following the original paper~\cite{tsai19}, we pass the input modalities through a 1D convolutional layer to embed the text series and the time series into the same latent space. Then we add positional encoding to each modality. Since we make predictions at $\m{X}_{EHR}$ frequency, we always obtain queries from $\m{X}_{EHR}$ and keys and values from $\m{X}_{EHR}$ or $\m{X}_{CN}$ in the self-attention and cross-attention layers, respectively. Figure~\ref{fig:architecture} shows the employed architecture.  

\paragraph{Tasks definition.}
Previous literature jointly using EHR data and clinical notes~\cite{khadanga19,chen20} relied on MIMIC-III data~\cite{johnson17}. Like other existing datasets, MIMIC-III consists of physiological measurements and information about laboratory tests. However, contrary to others, in MIMIC-III, a significant number of patients have clinical notes written by various medical personnel allowing for a multi-modal approach. Thus, for our experiments, we follow previous research by analyzing clinical notes' importance on \citet{harutyunyan19}'s benchmark tasks. In particular, we focus our work on three tasks. First, \textit{decompensation}, an hourly binary prediction task on whether the patient dies in the next 24 hours. Second, \textit{in-hospital mortality} (IHM), a similar task with a single binary prediction task after 48 hours of observed ICU data on whether the patient dies in the ICU. Finally, \textit{phenotyping}, a multi-class classification problem where one predicts which acute care conditions were present during a patient's stay out of 25 care conditions. The benchmark contains another task, hourly predicting the remaining length of stay. We do not consider this task due to its lower clinical relevance and the poor performance of all existing deep learning models on the task. We further discuss the data and its processing in Appendix~\ref{app:data}.

\paragraph{Hyperparameter tuning.} Increasing the number of heads or layers in transformers usually improves performance but also makes the models' analysis more complex. Since we already observed significant improvement over state-of-the-art uni-modal models using a single layer and head, we fixed these two hyperparameters to $1$ in all experiments. Doing so allowed us to preserve a more explainable model at the cost of a slight loss in performance. All other hyperparameters were selected using a grid search on the validation performance of \textit{decompensation}. For the convolutional layer, we use $64$ filters with a size equal to the input space. In the transformer, we use a latent space of size $64$ everywhere and a dropout rate of $0.2$. We trained all models with a batch size of $16$ patient stays, a learning rate of $1e^{-5}$, and Adam optimizer~\cite{kingma2014adam}.

\paragraph{Data processing.} As it is common for MIMIC-III benchmark data \cite{harutyunyan19}, we used forward imputation and standard scaling as pre-processing steps. More importantly, we mask out the last note in all tasks to avoid label leaks through direct death mentions. This is a simple way to reduce information leaking. Though, it is possible to tackle this more in-depth by filtering samples using keyword search as \cite{vanaken21}. As for the choice of hyperparameters, masking the last note might reduce absolute performance compared to previous research not following the same procedure~\cite{khadanga19,chen20}, but ensure a fairer analysis of the role of the clinical notes. 
 
\section{Experiments}
In this section, we first show that even with hard constraints on model complexity, jointly using EHR data and clinical notes outperforms existing uni-modal models on the considered tasks. We then perform an ablation study on note types and analyze cross-modal attention weights highlighting the importance of either ``nurse'' or ``radiology'' notes for cross-modal model decision-making. All results reported are from the test set and obtained over 5 random seeds. 

\paragraph{Overall performance.} As shown in Table \ref{tab:performance}, while only employing a single cross-attention layer with a single head and not attending to the last note, using a joint model surpasses previous literature's performance utilizing either EHR data~\citep{harutyunyan19} or clinical notes~\citep{khadanga19}. This observation confirms prior findings on the addition brought by clinical notes in patient monitoring \cite{khadanga19,chen20} and justifies the need to understand the underlying reason for such improvement. Interestingly, we also observe that our model performs on par with the best cross-modal model on both \textit{decompensation} and \textit{phenotyping} tasks despite our constraints.

\begin{table}[t]
\centering
\caption{Performances on MIMIC-III Benchmark~\citep{harutyunyan19} for uni-modal \textit{(top, middle)} and cross-modal models \textit{(bottom)}. Reported errors, if provided, are 95\% confidence intervals on the mean. }
\label{tab:performance}
\resizebox{\textwidth}{!}{
\begin{tabular}{@{}lcccccc@{}}
\toprule
Task & \multicolumn{2}{c}{Decompensation} & \multicolumn{2}{c}{IHM} & \multicolumn{2}{c@{}}{Phenotyping} \\
\cmidrule(lr){2-3} \cmidrule(lr){4-5} \cmidrule(l){6-7}
{Metric} & AUPRC & AUROC & AUPRC & AUROC & Macro-AUC & Micro-AUC  \\
\midrule
EHR-Only LSTM~\citep{harutyunyan19} & 33.3 $\pm$ 1.0 & 90.6 $\pm$ 0.3 & 51.5 $\pm$ 5.5 & 86.2 $\pm$ 2.2 & 77.6 $\pm$ 0.4 & 82.5 $\pm$ 0.3       \\
EHR-Only Transformer (\textbf{Ours})& 31.7 $\pm$ 0.4 & 90.3 $\pm$ 0.2 & 49.8 $\pm$ 0.9 & 85.5 $\pm$ 0.2 & 74.1 $\pm$ 0.1 & 79.9 $\pm$ 0.1         \\
\midrule
Text-only LSTM~\citep{khadanga19} & 8.1 $\pm$ 0.7 & 79.3 $\pm$ 0.7 & 30.3 $\pm$ 1.3 & 79.3 $\pm$ 0.4 & - & -        \\
Text-only Transformer (\textbf{Ours}) & 19.2 $\pm$ 0.6 & 83.3 $\pm$ 0.3 & 34.1 $\pm$ 0.6 & 77.9 $\pm$ 0.5 & 77.8 $\pm$ 0.2 & 82.7 $\pm$ 0.2         \\
\midrule
Cross-Modal LSTM~\citep{khadanga19} & 34.5 $\pm$ 0.7 & 90.7 $\pm$ 0.7 & 52.5 $\pm$ 1.3 & 86.5 $\pm$ 0.4 & - & -        \\
Cross-Modal Ensemble~\citep{chen20} & 40.4 $\pm$ n.a. & 92.0 $\pm$ n.a. & 58.2 $\pm$ n.a. & 88.6 $\pm$ n.a. & 82.9 $\pm$ n.a. & 87.0 $\pm$ n.a.        \\
Cross-Modal Transformer (\textbf{Ours}) & 39.7 $\pm$ 0.6 & 92.2 $\pm$ 0.2 & 52.7 $\pm$ 1.0 & 87.1 $\pm$ 0.6 & 82.6 $\pm$ 0.1 & 86.1 $\pm$ 0.1         \\
\bottomrule
\end{tabular}}
\end{table}

\begin{figure}[h]
    \centering
    \begin{subfigure}{0.49\textwidth}
        \centering
        \includegraphics[height=3.7cm,scale=1]{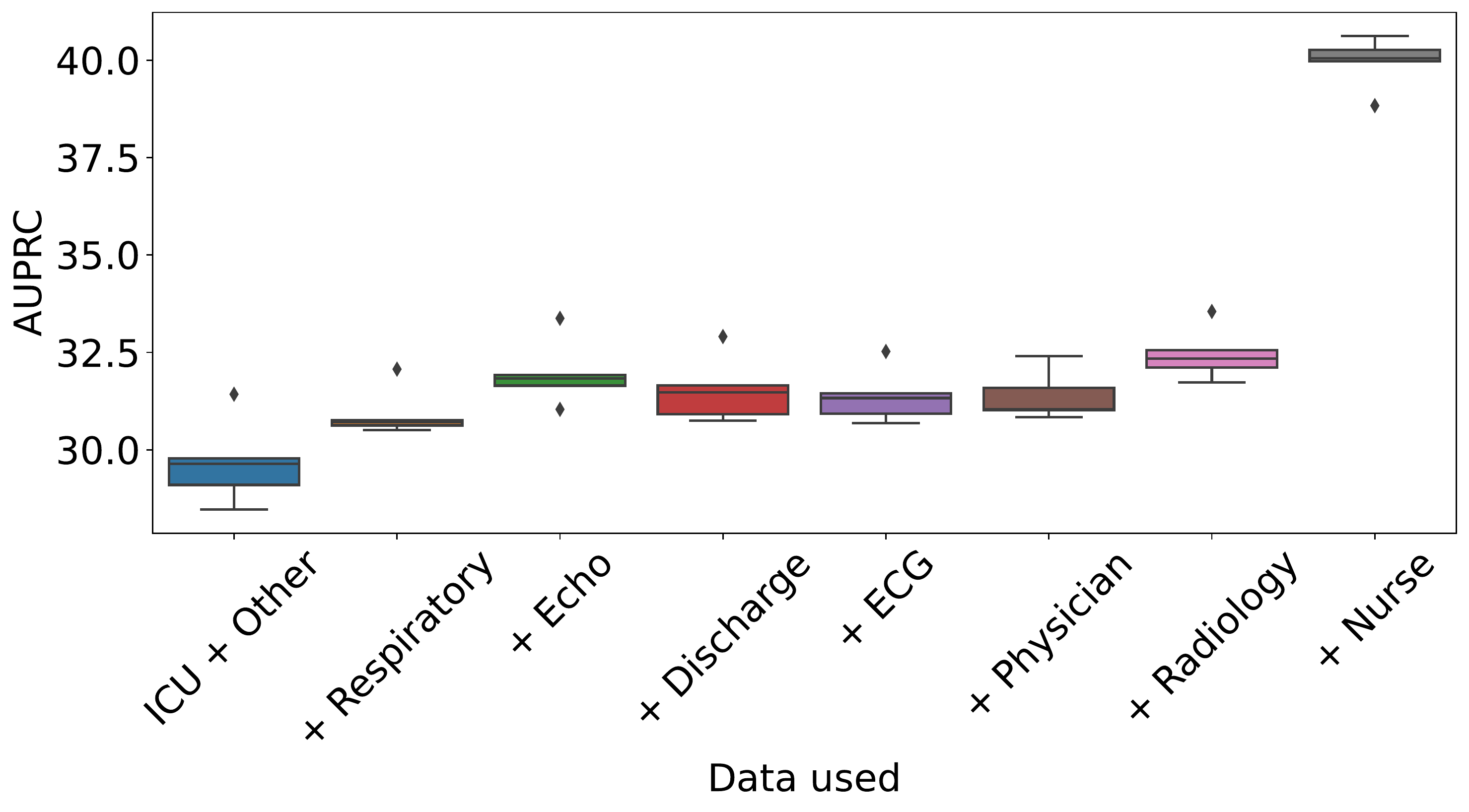}
        \label{fig:decompincreasing}
    \end{subfigure}
    \begin{subfigure}{0.49\textwidth}
        \centering
        \includegraphics[height=3.7cm,scale=1]{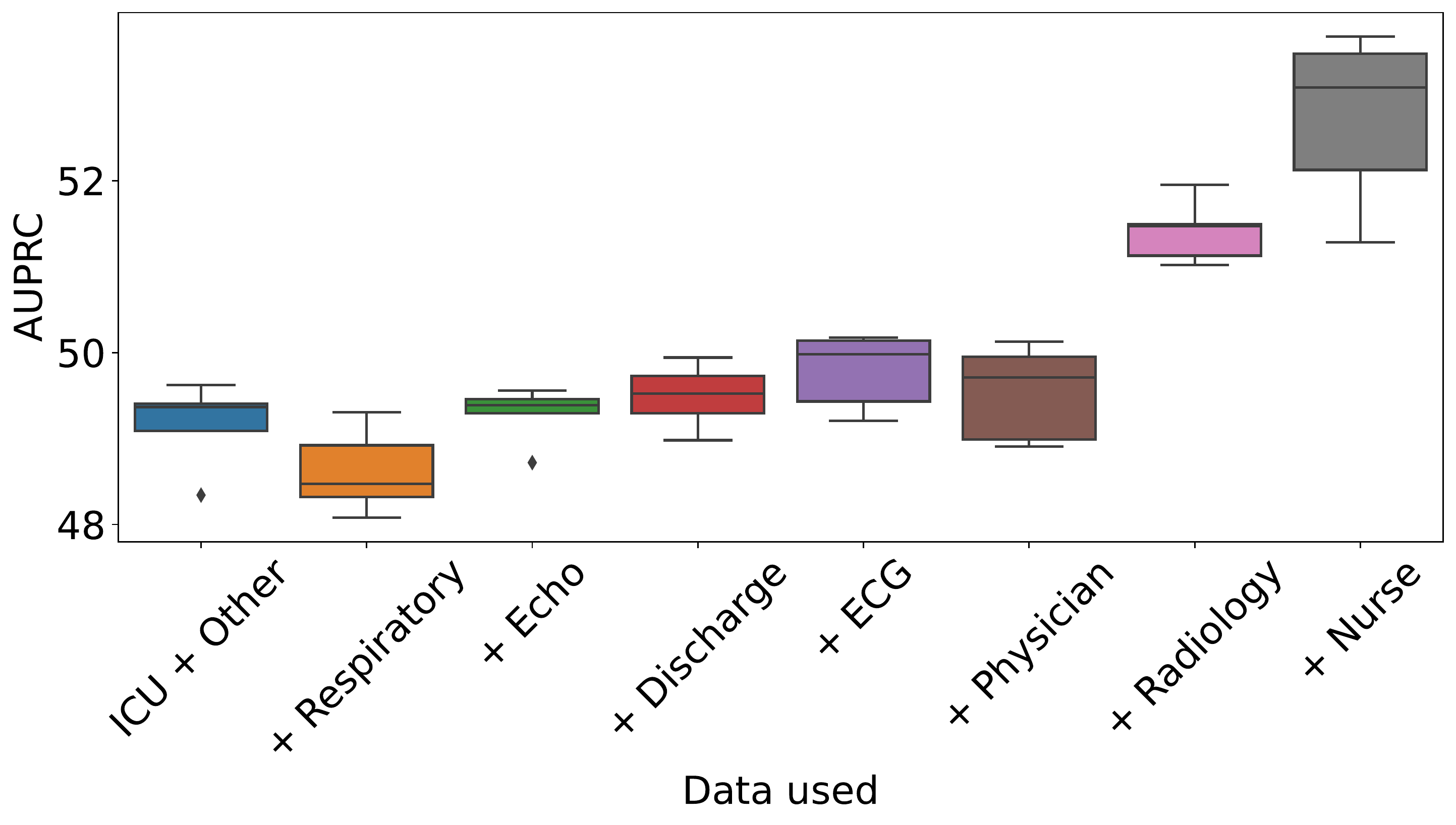}
        \label{fig:ihmincreasing}
    \end{subfigure}     
    \caption{Area under the precision-recall curve performance of the cross-modal transformer when adding different note types by increasing frequency. \textit{(left)} Decompensation.  \textit{(right)} In-hospital mortality.}
    \label{fig:increasingablation}
\end{figure}
\vspace{-1em}

\paragraph{Note type ablation.}
To understand which note types are important to the decision-making in our cross-modal model, we carry out ablation studies where we cumulatively add note types by increasing or by decreasing frequency (see Appendix~\ref{app:ablations}). Figure~\ref{fig:increasingablation} shows the ablation studies for the \textit{decompensation} and \textit{IHM} tasks by increasing frequency. We draw two conclusions. On the one hand, adding doctors' (``Physicians'') notes does not improve performance. This suggests that current representations obtained from the EHR data alone are sufficient. On the other hand, adding nurses' notes and radiology reports for \textit{IHM} significantly improves AUPRC on both tasks. Similar results can be found for \textit{phenotyping} in Appendix \ref{app:ablations}. These findings show that, when using clinical notes, deep learning models benefit from additional textual information rather than a possibly better representation of similar information in doctors' notes summarizing the patient's state.

\begin{figure}[h]
    \centering
    \begin{subfigure}[!t]{0.49\textwidth}
        \centering
        \includegraphics[height=3.6cm,scale=1]{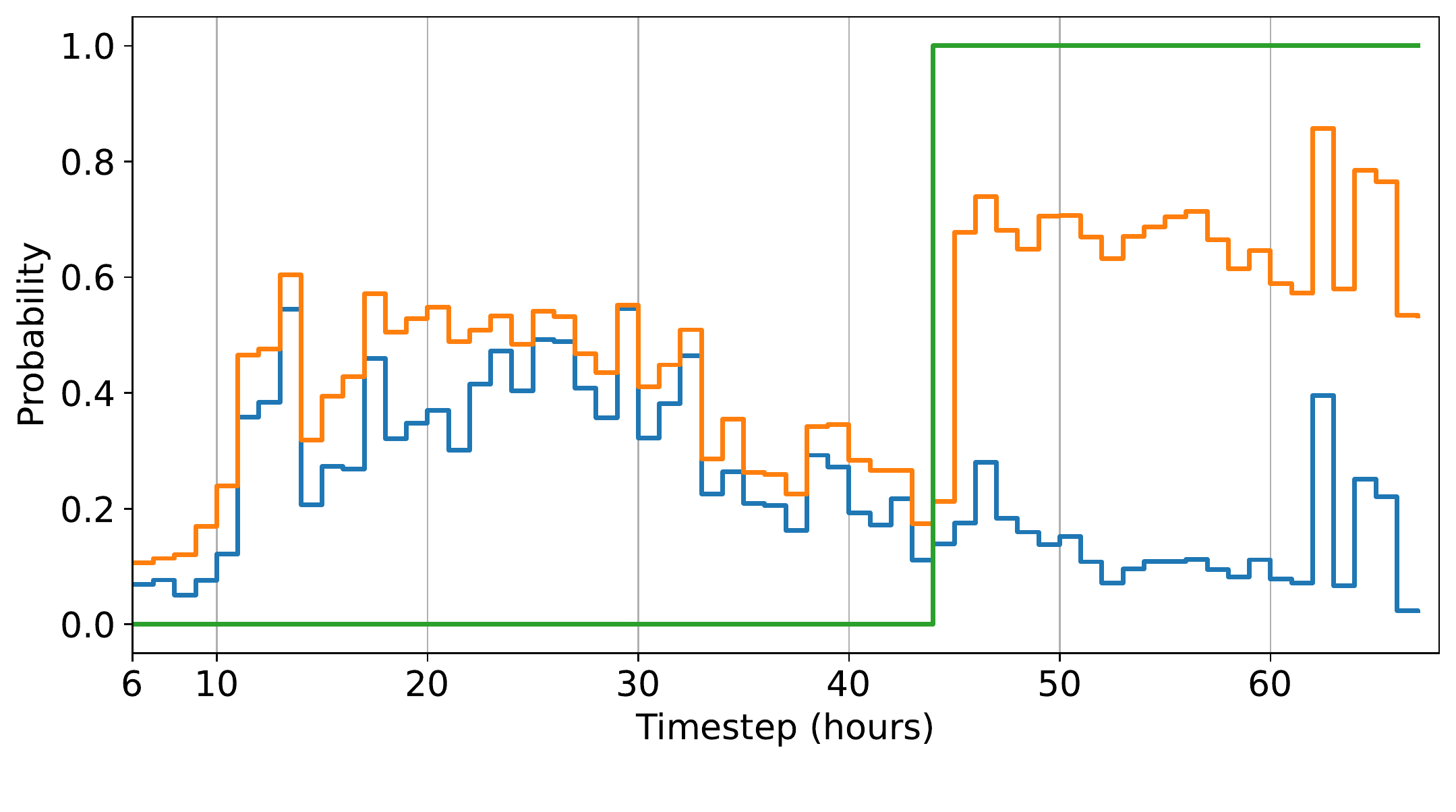}
    \end{subfigure}
    \begin{subfigure}[!t]{0.49\textwidth}
        \centering
        \includegraphics[height=3.6cm,scale=1]{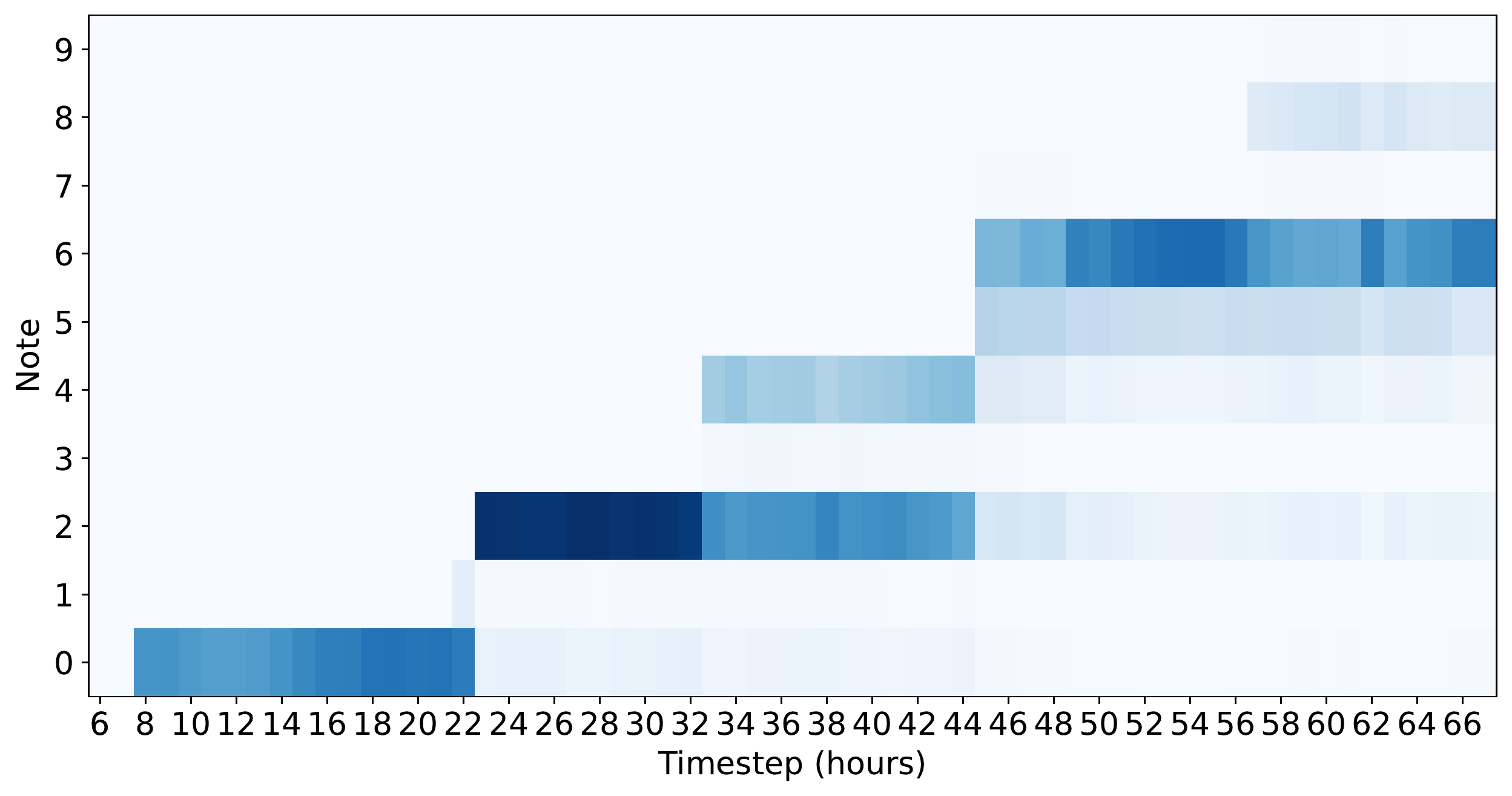}
    \end{subfigure}    
    \caption{Illustrative example of model's behaviour for decompensation. (\textit{left}) Predictions for EHR-only (blue) and cross-modal (orange) models over time compared to ground truth labels (green). (\textit{right}) Cross-attention over time for the cross-modal model where EHR timesteps (x-axis) attends to clinical notes (y-axis). Both figures represent patient 11555.}\vspace{-1.0em}
    \label{fig:interpretability11555}
\end{figure}

\paragraph{Cross-modal attention weights analysis.}
For all tasks, we observed that notes among the two most frequent types, nurses' and radiology notes, lead to the greatest improvements. Thus, one could argue that the note frequency explains the increase in performance over the note type itself. To answer this question, we perform an analysis of cross-modal attention weights. As shown with an illustrative example in Figure~\ref{fig:interpretability11555}, attention weights are very salient, making each EHR timestep attend to only a subset of notes. In addition, also shown in Figure \ref{fig:interpretability11555}, highly attended notes, such as the 7th note in the example, also impact the model's final prediction. Indeed, when it obtains this note at hour 45, the cross-modal model's predictions diverge from the EHR-only model ones. This high saliency of attention weights refutes the hypothesis that the nurses' and radiology notes are the most relevant types due to their high frequency. Indeed, in practice, only a subset of these notes are attended to independently of the notes type frequency. We provide additional examples to support this claim in Appendix~\ref{app:interpretability}.

\section{Conclusion}

In this paper, we aim to understand the underlying reason behind the significant improvement brought by jointly using EHR data and clinical notes in a multi-modal fashion in ICU-related tasks. To answer this question, we analyze the performance impact of the available note types in MIMIC-III and the cross-modal attention weights in our model. Based on our experimental results, we conclude that improvements yielded by adding clinical notes come from notes containing additional context on patient states, such as nurses' notes and radiology reports. We also find that doctors' notes, which often summarize EHR content, do not improve performance, suggesting that representations learned from EHR data alone preserve the input information. 
We believe these findings show how EHR data alone is only a narrow view of patient states in the ICU, motivating more data-centric approaches in the field.

\clearpage
\bibliographystyle{unsrtnat} 
\bibliography{refs}

\clearpage
\appendix

\section{Dataset}
\label{app:data}
\subsection{Pre-processing}

We first run the pre-processing step provided by \citet{harutyunyan19} for EHR data. Next, we extract all patients with notes. Echo, electrocardiogram (ECG), and discharge summary notes have no \emph{CHARTTIME}, the exact time at which the note is entered in the system. However, these notes might contain valuable medical information. Thus, we set their \emph{CHARTTIME} to the end of the day of their \emph{CHARTDATE}, which contains only the day on which the note was entered. It ensures no temporal leak of information. Following previous work~\cite{khadanga19}, we then exclude patients without any notes. From the benchmark task with 6281 ICU stays in the test set, we exclude 27 patients and end up with 6254 ICU stays. For some of these stays, only phenotyping task labels exist, hence, for the decompensation task, we end up with 6215 labeled ICU stays in the test set. We follow the same approach for the training set, validation set, and test set.

\subsection{Statistics}
For further details, we provide in Table~\ref{tab:testsetstats} the test set statistics for the \emph{decompensation} and \emph{IHM} tasks. In addition, Table \ref{tab:notestats} shows the note type counts in the training and test set. We ran the ablation studies according to the note type frequency in the training set.

\begin{table}[h]
\centering
\caption{Number of negative and positive labels for the IHM task and the decompensation task test sets.}
\label{tab:testsetstats}
     \begin{subtable}{0.45\textwidth}
        \centering
        \caption{Decompensation task test set}
        \label{tab:decomptestlabels}
        \begin{tabular}{@{}lr@{}}
        \toprule
        Label & Number of labels \\
        \midrule
            0 &           512862 \\
            1 &             9427 \\
        Total &           522289 \\
        \bottomrule
        \end{tabular}
    \end{subtable}
    \begin{subtable}{0.45\textwidth}
        \centering
        \caption{IHM task test set}
        \label{tab:ihmtestlabels}
        \begin{tabular}{@{}lr@{}}
        \toprule
        Label & Number of labels \\
        \midrule
            0 &             2860 \\
            1 &              365 \\
        Total &             3225 \\
        \bottomrule
        \end{tabular}
    \end{subtable}
\end{table}

\begin{table}[h]
\centering
\caption{Number of notes per note type in the test and training cohort.}
\label{tab:notestats}
     \begin{subtable}{0.45\textwidth}
        \centering
        \caption{Note type distribution in test set}
        \label{tab:notestatstest}
        \begin{tabular}{@{}lr@{}}
        \toprule
                    Note type &  Number of notes \\
        \midrule
                Nursing &            66206 \\
                    Radiology &            39744 \\
                      ECG &            16031 \\
                    Physician &            14261 \\
        Discharge summary &             6809 \\
                     Echo &             3762 \\
                  Respiratory &             3036 \\
                    Nutrition &              885 \\
                      General &              812 \\
               Rehab Services &              562 \\
                  Social Work &              263 \\
              Case Management &              102 \\
                     Pharmacy &               13 \\
                      Consult &               12 \\
        \bottomrule
        \end{tabular}
    \end{subtable}
    \begin{subtable}{0.45\textwidth}
        \centering
        \caption{Note type distribution in training set}
        \label{tab:notestatstrain}
        \begin{tabular}{@{}lr@{}}
        \toprule
                    Note type &  Number of notes \\
        \midrule
                Nursing &           316015 \\
                    Radiology &           183490 \\
                    Physician &            74089 \\
                      ECG &            73227 \\
        Discharge summary &            31561 \\
                     Echo &            17298 \\
                  Respiratory &            16247 \\
                    Nutrition &             4777 \\
                      General &             4376 \\
               Rehab Services &             2744 \\
                  Social Work &             1400 \\
              Case Management &              535 \\
                     Pharmacy &               50 \\
                      Consult &               43 \\
        \bottomrule
        \end{tabular}
    \end{subtable}
\end{table}

\section{Ablation Studies}
\label{app:ablations}

In this section, we provide further results regarding notes type ablations.

\subsection{Increasing Frequency}
We first summarize results from the ablation by increasing frequency from Figure~\ref{fig:increasingablation} in Table~\ref{tab:ablation}. We find, as previously mentioned, that for all tasks, nurses' or radiology notes contribute to improved performance. For the phenotyping task, we see that also other note types help. It may be because these types contain information for specific illnesses not necessarily present in the other notes, thus also from some additional context on patients' states. Nonetheless, the main contribution to performance remains from radiology reports.

\begin{table}
\centering
\caption{Note type ablation study results when adding clinical notes by increasing frequency.}
\label{tab:ablation}
\begin{tabular}{@{}lrrrrrr@{}}
\toprule
Task & \multicolumn{2}{c}{Decompensation} & \multicolumn{2}{c}{IHM} & \multicolumn{2}{c@{}}{Phenotyping} \\
\cmidrule(lr){2-3} \cmidrule(lr){4-5} \cmidrule(l){6-7}
{} & AUPRC & AUROC & AUPRC & AUROC & \multicolumn{2}{c@{}}{AUROC} \\
Model & {} & {} & {} & {} & \multicolumn{1}{c}{Macro} & \multicolumn{1}{c}{Micro} \\

\midrule
        ICU + Other & 29.7 $\pm$ 1.1 & 90.0 $\pm$ 0.1 & 49.2 $\pm$ 0.5 & 85.4 $\pm$ 0.2 & 74.2 $\pm$ 0.3 &  80.0 $\pm$ 0.2 \\
      + Respiratory & 30.9 $\pm$ 0.6 & 90.1 $\pm$ 0.1 & 48.6 $\pm$ 0.5 & 85.1 $\pm$ 0.2 & 74.2 $\pm$ 0.2 &  80.0 $\pm$ 0.1 \\
             + Echo & 32.0 $\pm$ 0.9 & 90.2 $\pm$ 0.1 & 49.3 $\pm$ 0.3 & 84.9 $\pm$ 0.2 & 74.8 $\pm$ 0.1 &  80.4 $\pm$ 0.1 \\
        + Discharge & 31.5 $\pm$ 0.9 & 90.1 $\pm$ 0.1 & 49.5 $\pm$ 0.4 & 84.9 $\pm$ 0.2 & 76.0 $\pm$ 0.1 &  81.3 $\pm$ 0.1 \\
              + ECG & 31.4 $\pm$ 0.7 & 90.1 $\pm$ 0.1 & 49.8 $\pm$ 0.4 & 85.8 $\pm$ 0.2 & 78.0 $\pm$ 0.2 &  82.8 $\pm$ 0.1 \\
        + Physician & 31.4 $\pm$ 0.6 & 90.0 $\pm$ 0.2 & 49.5 $\pm$ 0.6 & 85.7 $\pm$ 0.2 & 78.6 $\pm$ 0.2 &  83.2 $\pm$ 0.1 \\
        + Radiology & 32.5 $\pm$ 0.7 & 90.6 $\pm$ 0.2 & 51.4 $\pm$ 0.4 & 86.4 $\pm$ 0.4 & 81.4 $\pm$ 0.2 &  85.2 $\pm$ 0.1 \\
            + Nurse & 39.9 $\pm$ 0.7 & 92.3 $\pm$ 0.2 & 52.7 $\pm$ 1.0 & 87.1 $\pm$ 0.6 & 82.6 $\pm$ 0.1 &  86.1 $\pm$ 0.1 \\
\bottomrule
\end{tabular}
\end{table}

\begin{figure}[h]
    \centering
    \begin{subfigure}{0.49\textwidth}
        \centering
        \includegraphics[height=3.8cm,scale=1]{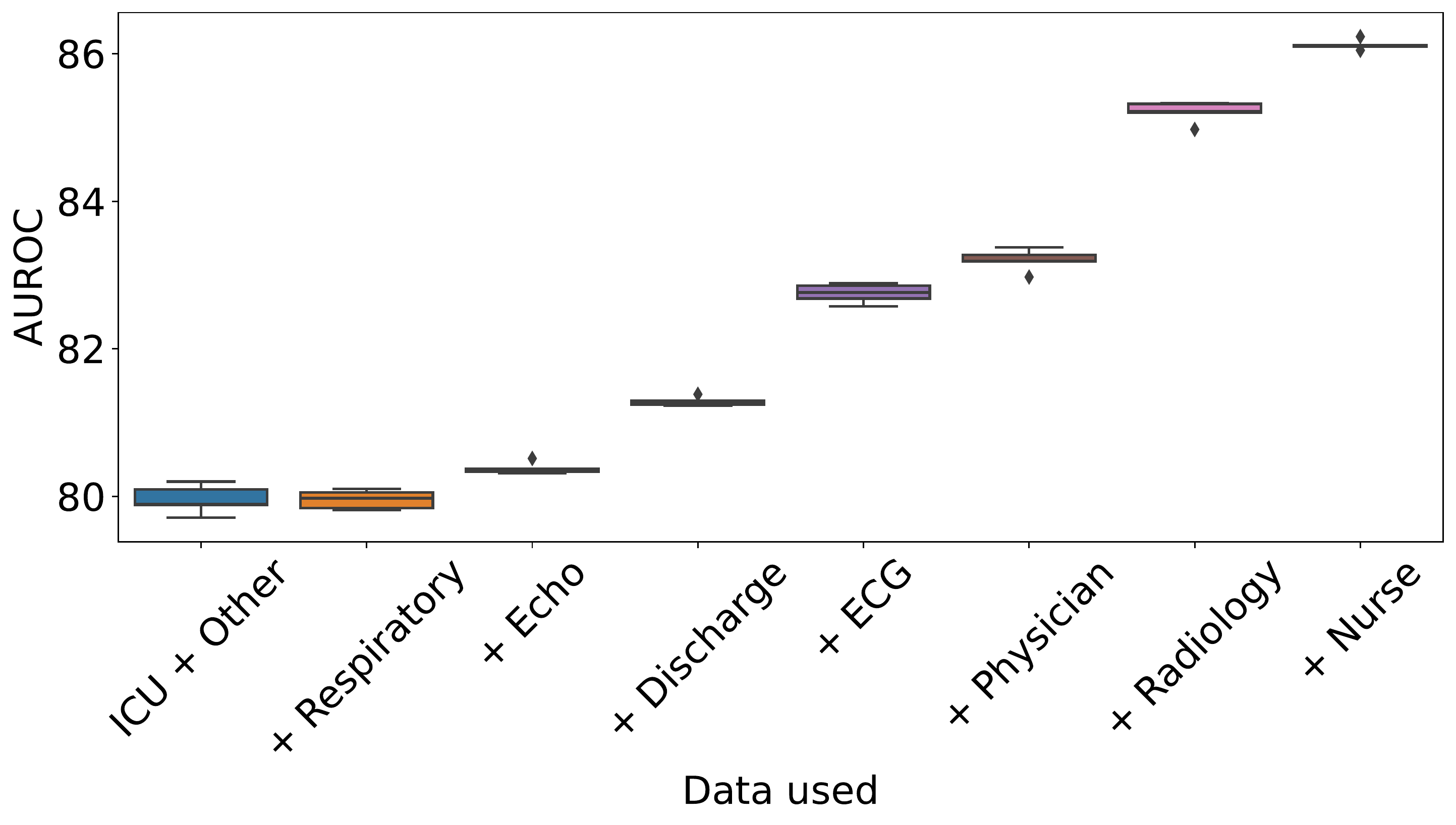}
\end{subfigure}
    \begin{subfigure}{0.49\textwidth}
        \centering
        \includegraphics[height=3.8cm,scale=1]{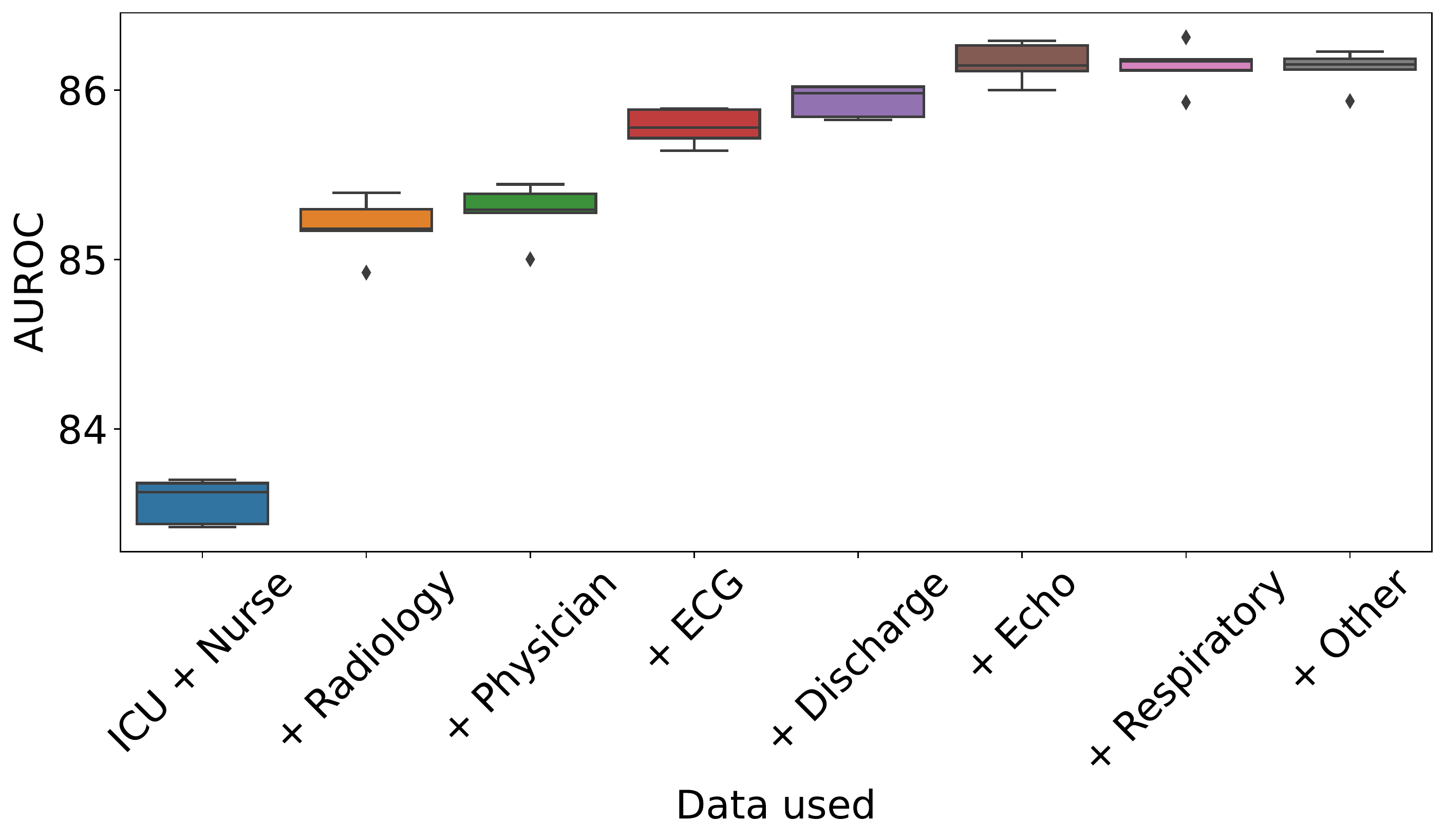}
\end{subfigure}
    \caption{Micro-averaged AUROC note type ablation study results for the phenotyping task. \textit{(left)} By increasing note type frequency. \textit{(right)} by decreasing note type frequency.}
    \label{fig:ablationspheno}
\end{figure}

\subsection{Decreasing Frequency}

\begin{figure}[h]
    \centering
    \begin{subfigure}{0.49\textwidth}
        \centering
        \includegraphics[height=3.8cm,scale=1]{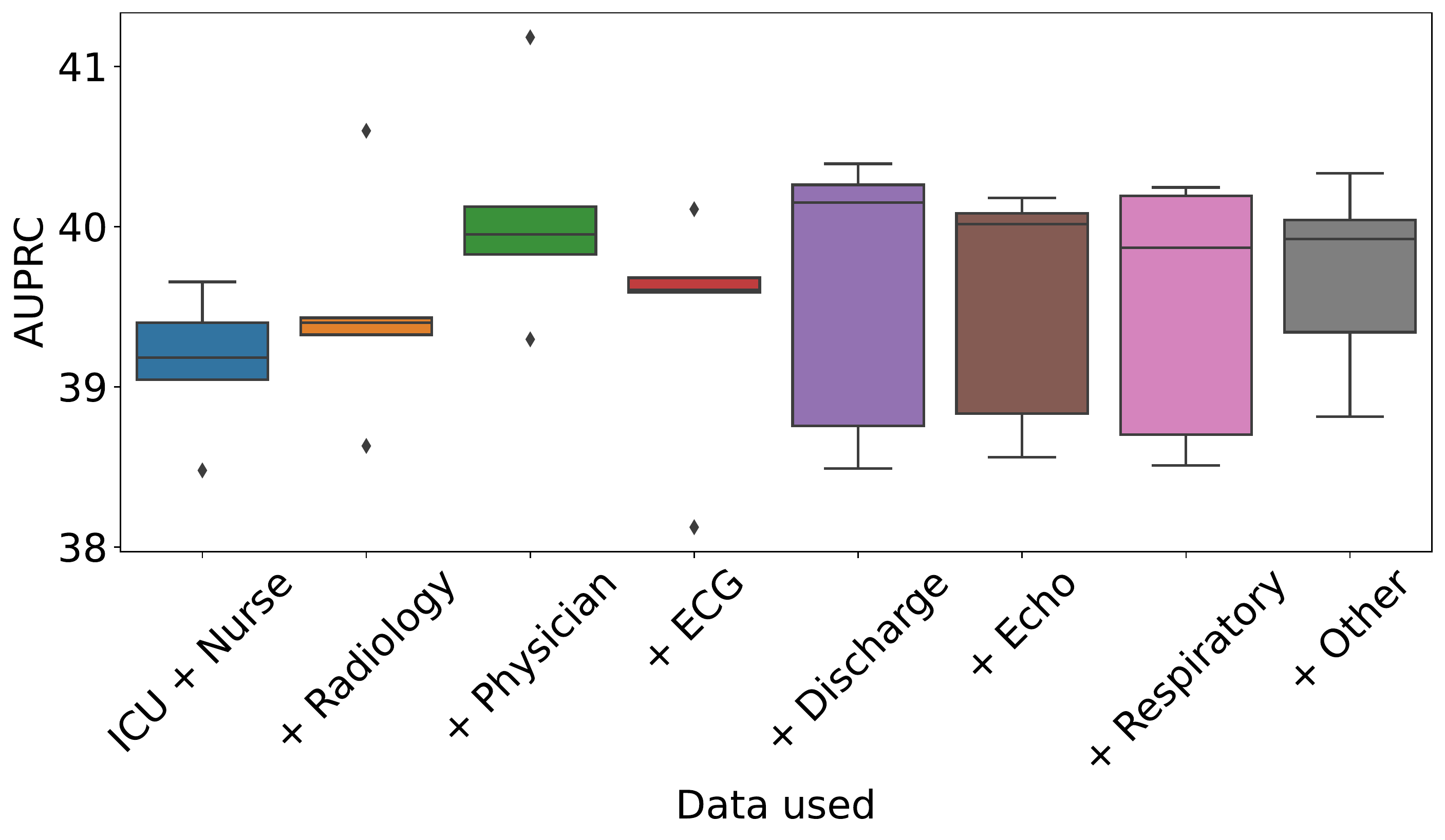}
\end{subfigure}
    \begin{subfigure}{0.49\textwidth}
        \centering
        \includegraphics[height=3.8cm,scale=1]{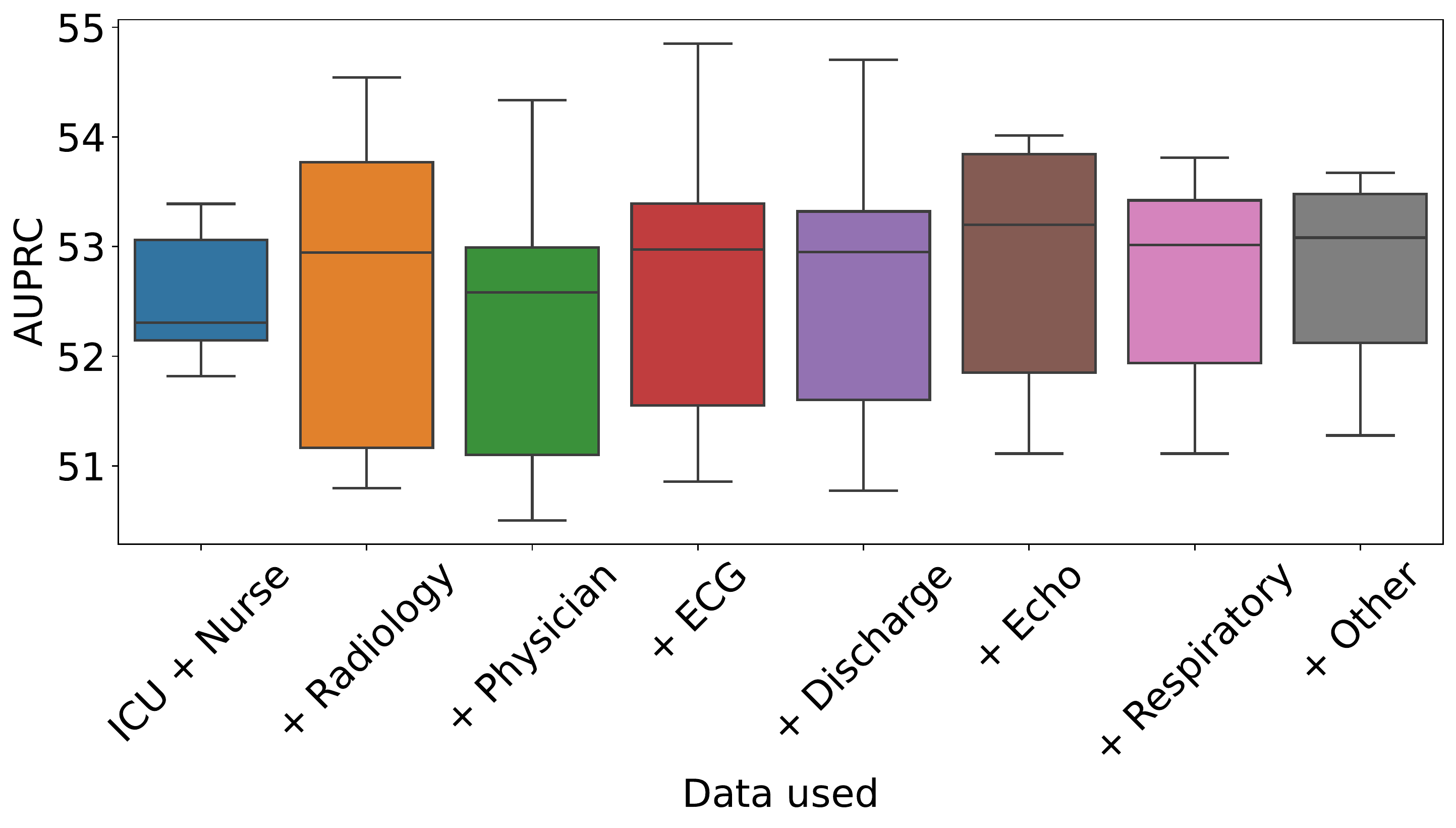}
\end{subfigure}
    \caption{Note type ablation study results by adding types by decreasing frequency. \textit{(left)} Decompensation. \textit{(right)} In-hospital mortality.}
    \label{fig:decreasingablation}
\end{figure}

We also provide an ablation study on note types by decreasing frequency, as a sanity check, in Figure~\ref{fig:decreasingablation} and Table~\ref{tab:ablationdecr}. We draw the same conclusions on note types importance.

\begin{table}
\centering
\caption{Note type ablation study results when adding clinical notes by decreasing frequency.}
\label{tab:ablationdecr}
\begin{tabular}{@{}lrrrrrr@{}}
\toprule
Task & \multicolumn{2}{c}{Decompensation} & \multicolumn{2}{c}{IHM} & \multicolumn{2}{c@{}}{Phenotyping} \\
\cmidrule(lr){2-3} \cmidrule(lr){4-5} \cmidrule(l){6-7}
{} & AUPRC & AUROC & AUPRC & AUROC & \multicolumn{2}{c@{}}{AUROC} \\
Model & {} & {} & {} & {} & \multicolumn{1}{c}{Macro} & \multicolumn{1}{c}{Micro} \\

\midrule
        ICU + Nurse & 39.2 $\pm$ 0.4 & 92.3 $\pm$ 0.1 & 52.5 $\pm$ 0.7 & 86.6 $\pm$ 0.2 & 79.2 $\pm$ 0.2 & 83.6 $\pm$ 0.1 \\
        + Radiology & 39.5 $\pm$ 0.7 & 92.4 $\pm$ 0.1 & 52.6 $\pm$ 1.6 & 86.8 $\pm$ 0.9 & 81.4 $\pm$ 0.2 & 85.2 $\pm$ 0.2 \\
        + Physician & 40.1 $\pm$ 0.7 & 92.4 $\pm$ 0.2 & 52.3 $\pm$ 1.5 & 86.8 $\pm$ 0.8 & 81.5 $\pm$ 0.2 & 85.3 $\pm$ 0.2 \\
              + ECG & 39.4 $\pm$ 0.8 & 92.3 $\pm$ 0.2 & 52.7 $\pm$ 1.6 & 87.1 $\pm$ 0.8 & 82.2 $\pm$ 0.1 & 85.8 $\pm$ 0.1 \\
        + Discharge & 39.6 $\pm$ 0.9 & 92.3 $\pm$ 0.3 & 52.7 $\pm$ 1.5 & 87.1 $\pm$ 0.8 & 82.3 $\pm$ 0.1 & 85.9 $\pm$ 0.1 \\
             + Echo & 39.5 $\pm$ 0.8 & 92.2 $\pm$ 0.2 & 52.8 $\pm$ 1.3 & 87.1 $\pm$ 0.7 & 82.7 $\pm$ 0.1 & 86.2 $\pm$ 0.1 \\
      + Respiratory & 39.5 $\pm$ 0.8 & 92.2 $\pm$ 0.2 & 52.7 $\pm$ 1.1 & 87.1 $\pm$ 0.7 & 82.6 $\pm$ 0.2 & 86.1 $\pm$ 0.1 \\
            + Other & 39.7 $\pm$ 0.6 & 92.2 $\pm$ 0.2 & 52.7 $\pm$ 1.0 & 87.1 $\pm$ 0.6 & 82.6 $\pm$ 0.1 & 86.1 $\pm$ 0.1 \\
\bottomrule
\end{tabular}
\end{table}

\section{Interpretability results}
\label{app:interpretability}
In this section, we describe our attempt to analyze notes importance more in-depth through attention rollout and chunk level representations. We also provide a similar analysis for another illustrative example. 

\subsection{Deeper analysis on main manuscript example}
\begin{figure}[h]
    \centering
    \includegraphics[height=4cm,scale=1]{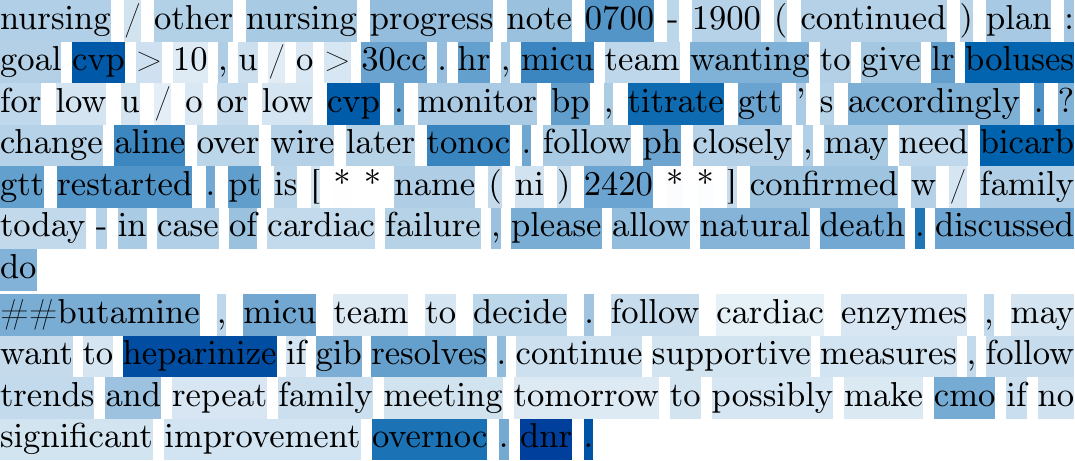}
    \caption{Attention rollout of Clinical BERT on 7th note from patient 11555}
    \label{fig:rollout11555_n6}
\end{figure}

In the example from the main manuscript, we found that the 7th note was strongly attended to in Figure \ref{fig:interpretability11555}. To further understand, what lead this note to significantly change the model prediction toward a imminent death, we ran attention rollout~\cite{abnar20} and summed over the words~\cite{clark19} (see Figure~\ref{fig:rollout11555_n6}). Additionally, we normalized by the number of words in each chunk when a note does not fit a single Clinical BERT pass (i.e., it is longer than 128 tokens).
One interesting observation is how \emph{dnr} is significant in the overall note representation. Indeed this stands for ``do no resucitate'', indicating a clear change of treatment in case of cardiac or respiratory arrest and thus a more probable death.

\begin{figure}[h]
    \centering
    \begin{subfigure}{0.8\textwidth}
        \centering
        \includegraphics{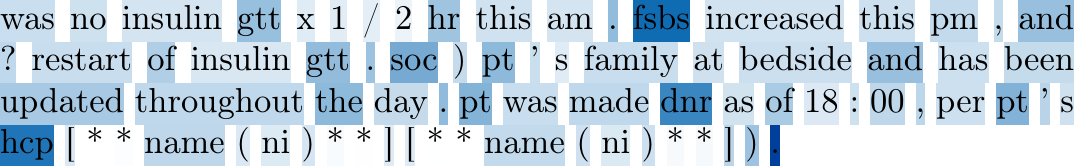}
        \caption{Attention rollout for 12th chunk}
        \label{fig:decompex1chunk11}
    \end{subfigure}
    \begin{subfigure}{0.8\textwidth}
        \centering
        \includegraphics{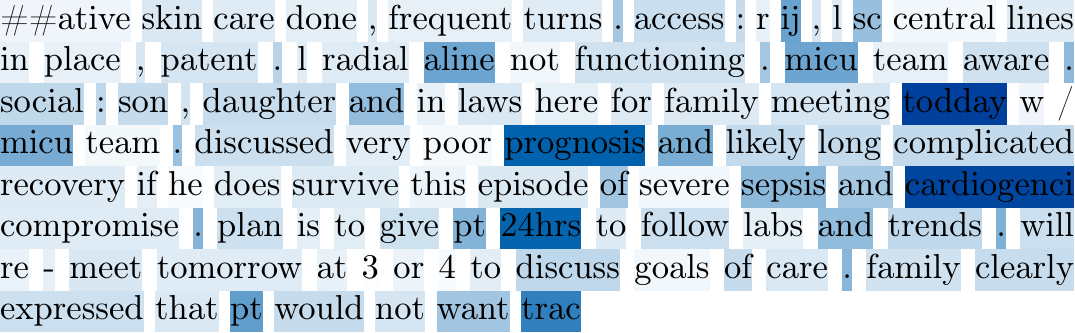}
        \caption{Attention rollout for 25th chunk}
        \label{fig:decompex1chunk24}
    \end{subfigure}    
    \begin{subfigure}{0.8\textwidth}
        \centering
        \includegraphics{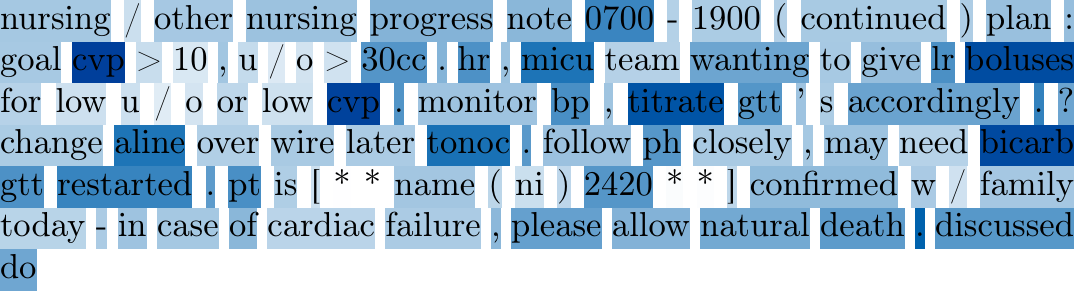}
        \caption{Attention rollout for 27th chunk}
        \label{fig:decompex1chunk26}
    \end{subfigure}    
    \caption{Attention rollout over Clinical BERT from nurse's note chunks for the patient with ID 11555}
    \label{fig:rollout11555_chunks}
\end{figure}

\paragraph{Chunk level representation of notes.} In our model, if a note is longer than 128 tokens on which Clinical BERT was trained, we passed the chunks separately through Clinical BERT and averaged the global token embedding. For an additional level of interpretability, we use the same model architecture and train using all embedded chunks of the same note with the same time appended. For the identical patient, we see that chunk 12 is attended early on, and the 25th and 27th chunks are the strongest attended to when the prediction gap increases. Figure \ref{fig:rollout11555_chunks} shows the attention rollout results on the chunks separately. Once again in the most attended chunk (see Figure~\ref{fig:decompex1chunk26}), we see the nurses mentions to ``please allow natural death``. 

\begin{figure}[h]
    \centering
    \begin{subfigure}[!t]{0.49\textwidth}
        \centering
        \includegraphics[height=3.8cm,scale=1]{figures/preds_11555_1.pdf}
\end{subfigure}
    \begin{subfigure}[!t]{0.49\textwidth}
        \centering
        \includegraphics[height=3.4cm,scale=1]{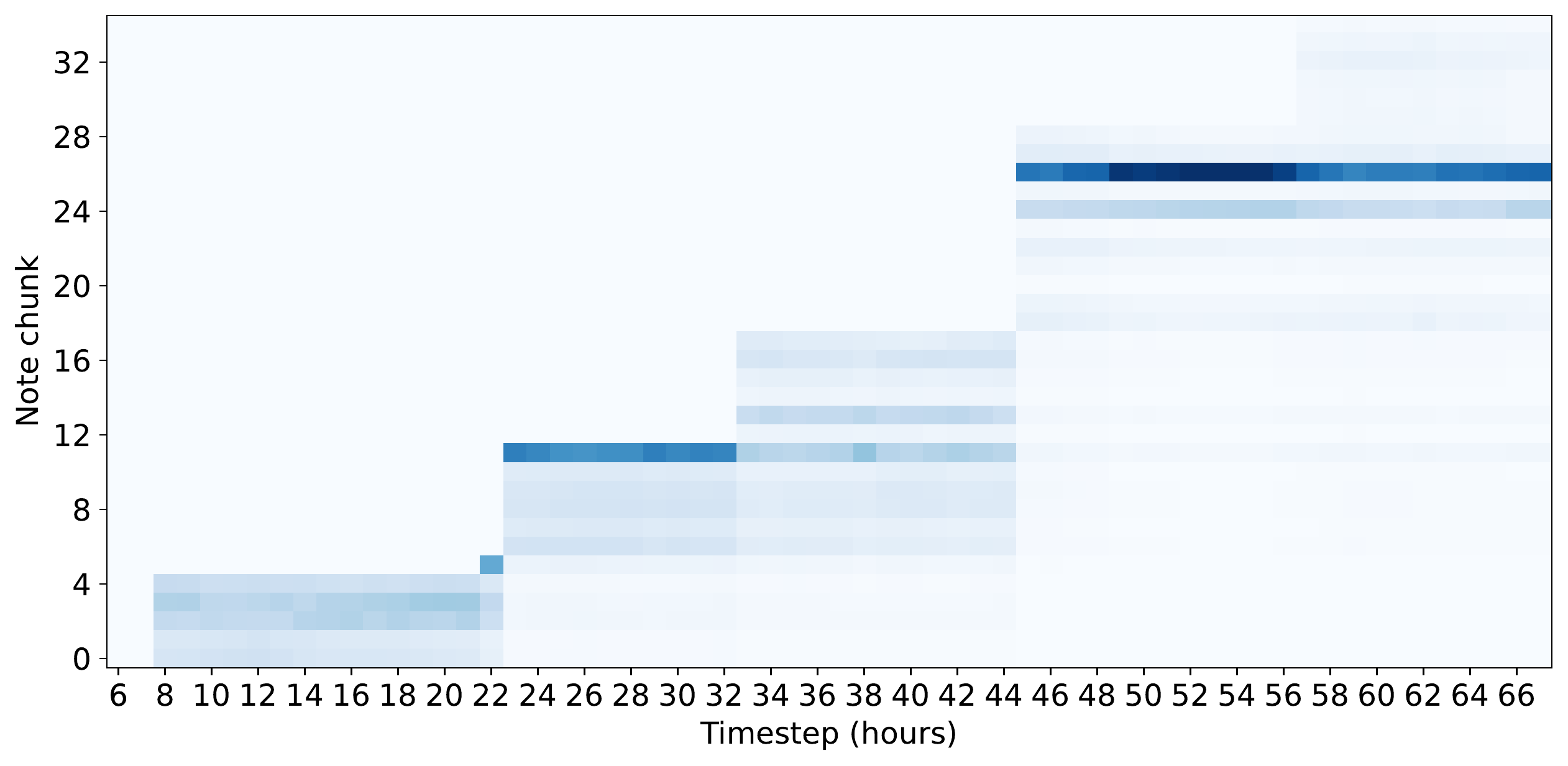}
\end{subfigure}    
\caption{Illustrative example of chunk model's behaviour for decompensation. (\textit{left}) Predictions for EHR-only (blue) and cross-modal (orange) models over time compared to ground truth labels (green). (\textit{right}) Chunk cross-attention over time for the cross-modal model where EHR timesteps (x-axis) attend to clinical notes (y-axis). Both figures represent patient 11555.}
    \label{fig:interpretability11555_chunk}
\end{figure}

\subsection{Other Illustrative Example}
\begin{figure}[h]
    \centering
    \begin{subfigure}[!t]{0.49\textwidth}
        \centering
        \includegraphics[height=3.8cm,scale=1]{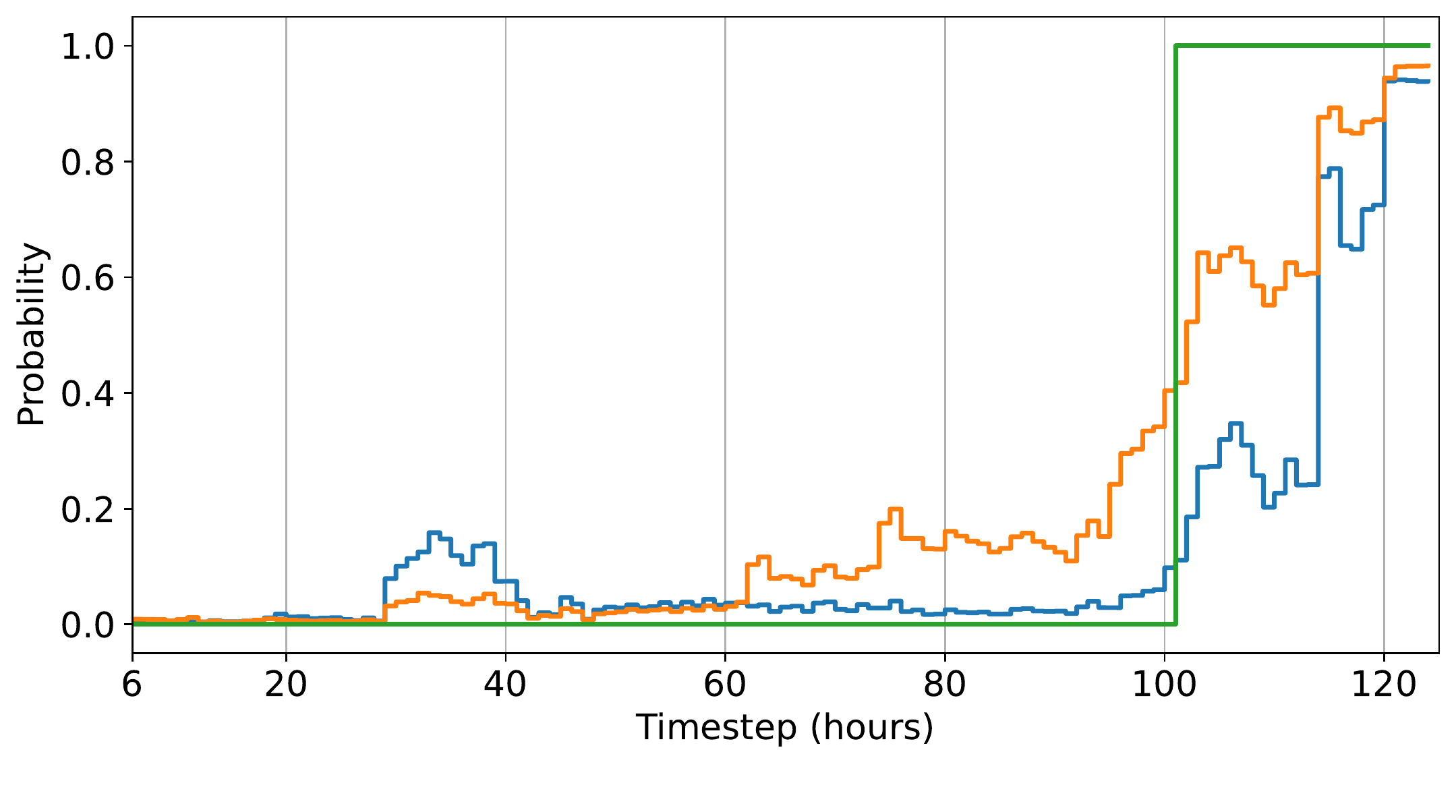}
\end{subfigure}
    \begin{subfigure}[!t]{0.49\textwidth}
        \centering
        \includegraphics[height=3.6cm,scale=1]{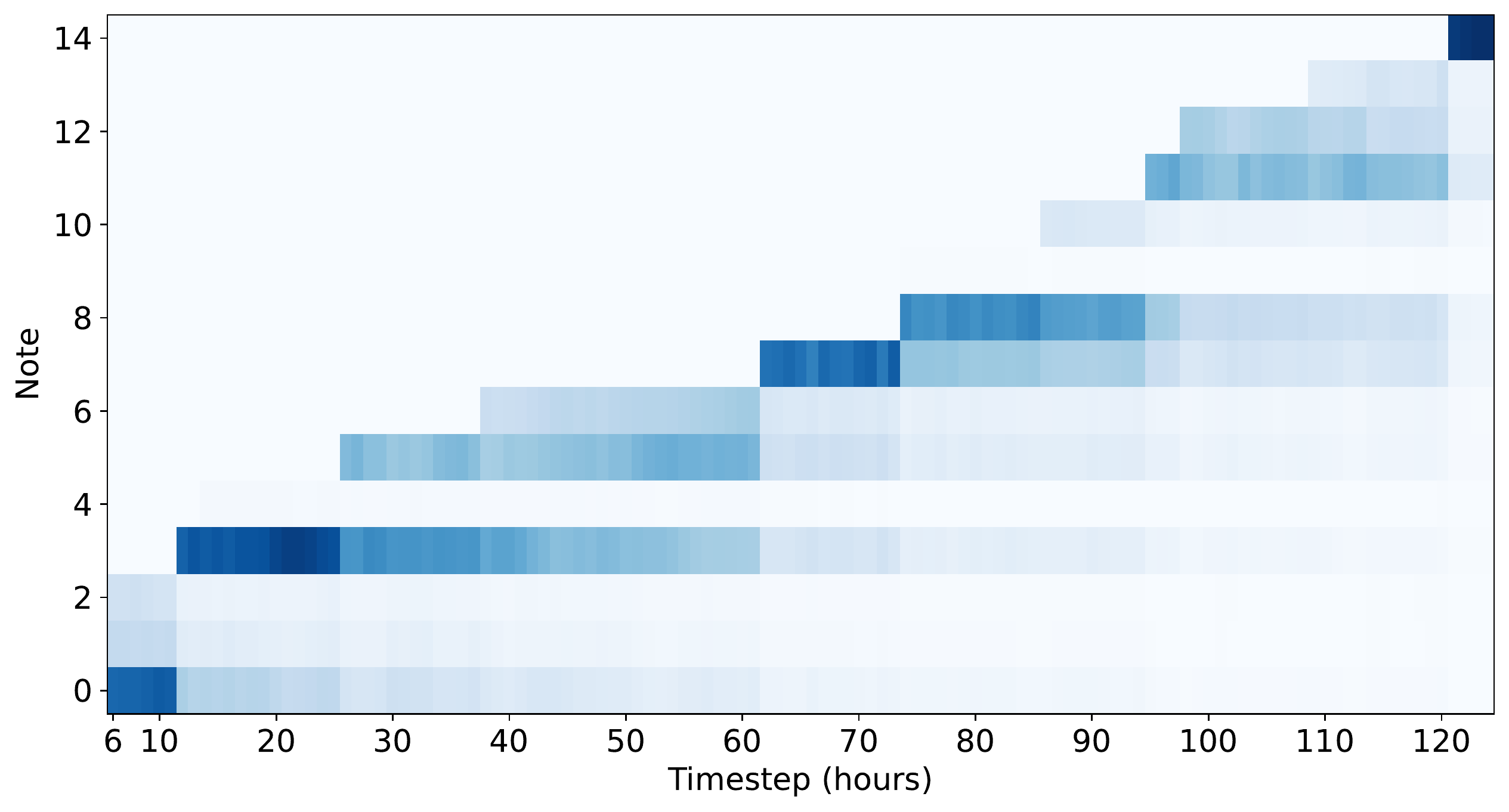}
\end{subfigure}    
\caption{Illustrative example of model's behaviour for decompensation. (\textit{left}) Predictions for EHR-only (blue) and cross-modal (orange) models over time compared to ground truth labels (green). (\textit{right}) Cross-attention over time for the cross-modal model where EHR timesteps (x-axis) attend to clinical notes (y-axis). Both figures represent patient 12767.}
    \label{fig:interpretability12767}
\end{figure}

To further illustrate our conclusions, we also present a second example corresponding to the patient with ID 12767. Figure~\ref{fig:interpretability12767} shows the predictions and the corresponding cross-attention from ICU time series to text series over time. We observe that at the 63rd timestep when the multi-modal transformer's prediction deviates from the EHR-only transformer's prediction it attends to the newly available 8th note.

\begin{figure}[h]
    \centering
    \includegraphics[width=0.8\textwidth,scale=1]{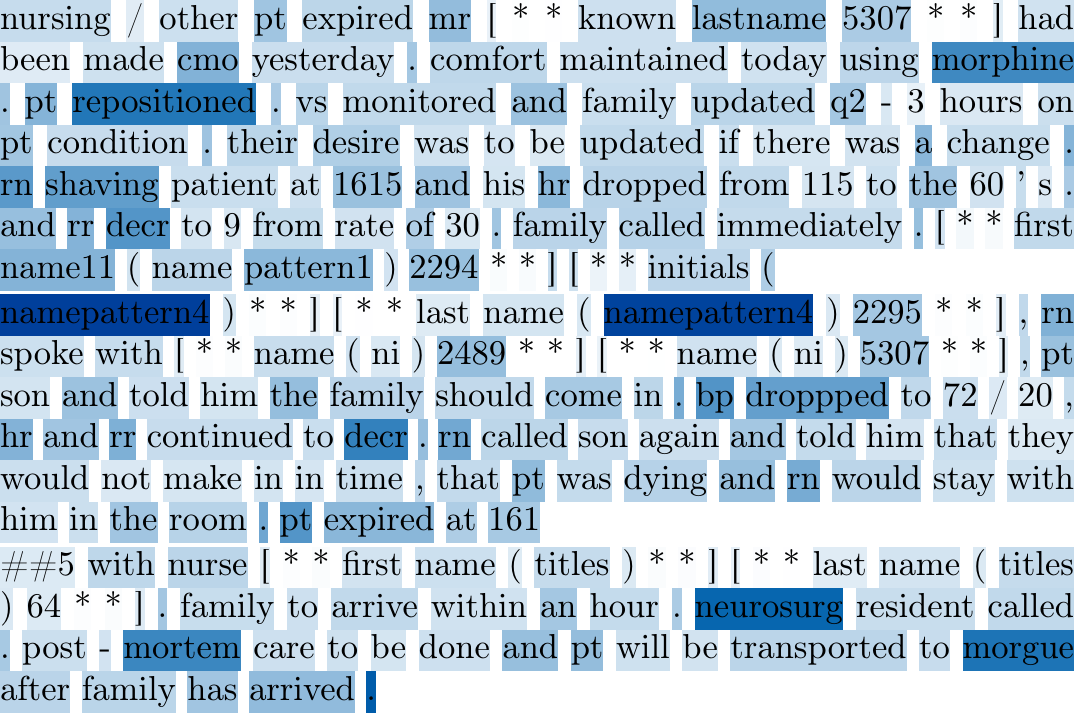}
    \caption{Attention rollout of Clinical BERT of last note from the patient with ID 12767.}
    \label{fig:decompex2note14}
\end{figure}

 As in the previous example, we compute the attention rollout from this particular note in Figure~\ref{fig:interpretability12767}. This short note states that the patient is most likely to be made DNR/DNI. We see that besides the DNR status, attention is seemingly strongly placed also on \emph{unresponsive} and \emph{tylenol}. Interestingly, we observe that the last note is highly attended to. Indeed, we see that the note already discusses the patient's death, highlighting the existing leakage of information. Hence, it is clear why the model is above 90\% certain that the patient will die within the next 24 hours. We show this last note attention rollout in Figure~\ref{fig:decompex2note14}. Interestingly, not necessarily \emph{expired} is focused in contrast to \emph{mortem}. We believe this is because the former is a clinical term, thus, more frequent in the corpus used to train clinical BERT.

\begin{figure}[h]
    \centering
    \includegraphics[width=0.8\textwidth,scale=1]{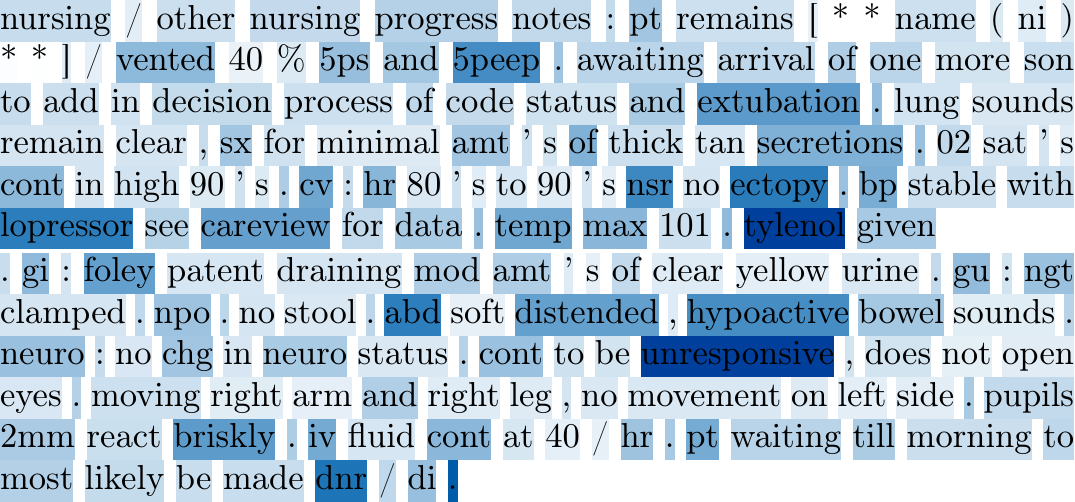}
    \caption{Attention rollout of Clinical BERT of 8th note from the patient with ID 12767}
    \label{fig:decompex2note7}
\end{figure}

\begin{figure}
    \centering
    \includegraphics[width=0.8\textwidth,scale=1]{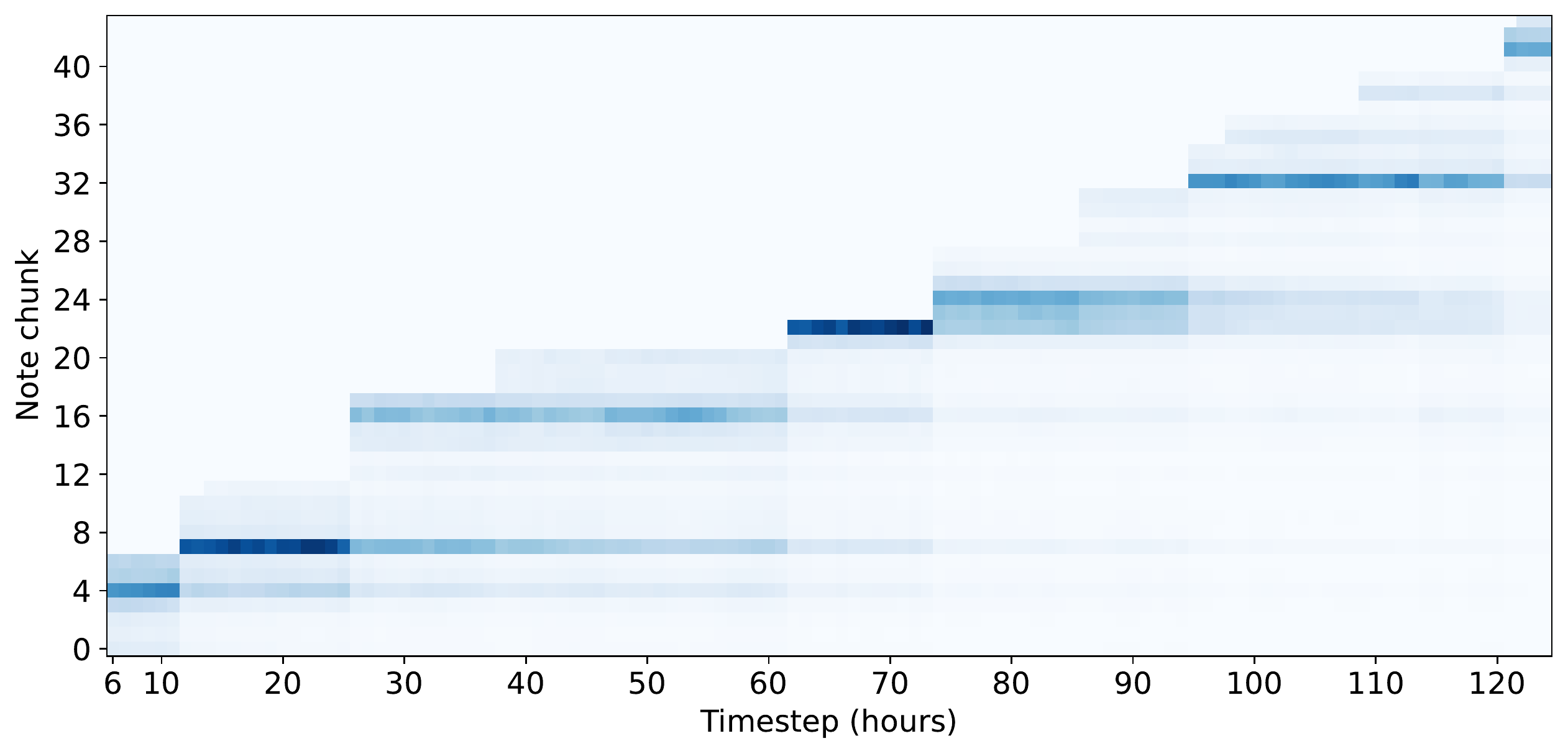}
\caption{Chunk cross-attention over time for the cross-modal model where EHR timesteps (x-axis) attends to clinical notes (y-axis) for the patient with ID 12767}
    \label{fig:crossatt-chunk12767}
\end{figure}

\begin{figure}
    \centering
    \begin{subfigure}{\textwidth}
        \centering
        \includegraphics{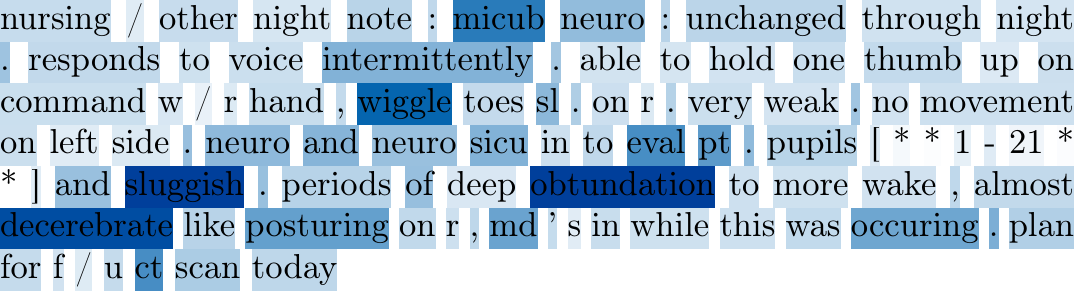}
        \caption{Attention rollout for 8th chunk}
        \label{fig:decompex2chunk7}
    \end{subfigure}
    \begin{subfigure}{\textwidth}
        \centering
        \includegraphics{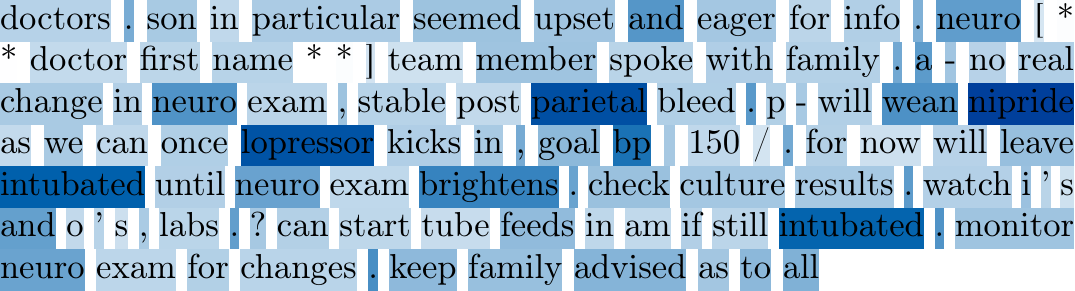}
        \caption{Attention rollout for 17th chunk}
        \label{fig:decompex2chunk16}
    \end{subfigure}    
    \begin{subfigure}{\textwidth}
        \centering
        \includegraphics{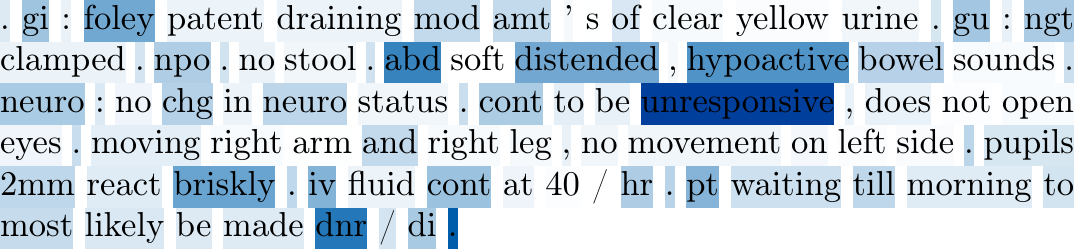}
        \caption{Attention rollout for 23rd chunk}
        \label{fig:decompex2chunk22}
    \end{subfigure}    
    \begin{subfigure}{\textwidth}
        \centering
        \includegraphics{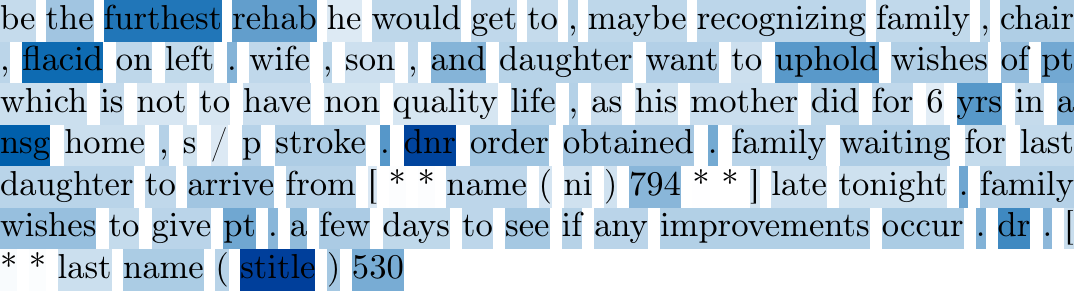}
        \caption{Attention rollout for 25th chunk}
        \label{fig:decompex2chunk24}
    \end{subfigure}    
    \begin{subfigure}{\textwidth}
        \centering
        \includegraphics{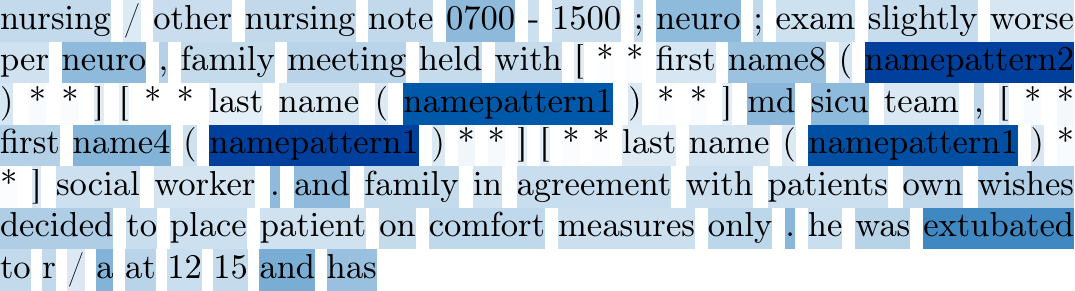}
        \caption{Attention rollout for 33rd chunk}
        \label{fig:decompex2chunk32}
    \end{subfigure}    
    \caption{Attention rollout over Clinical BERT from nurse's note chunks for the patient with ID 12767}
    \label{fig:rollout12767_chunks}
\end{figure}

As for the previous examples, we also explore the chunks of the notes directly. Figure \ref{fig:crossatt-chunk12767} shows the ICU to nurse note chunks cross-attention over time. For this patient the 8th, 17th, 23rd, 25th, and 33rd chunks are strongly attended to over time. In Figure \ref{fig:rollout12767_chunks} we present the corresponding attention rollout results. In chunk 7, we find medically relevant observations like \emph{obtundation} or \emph{decerebrate like posturing} attended to. Obtundation describes the status when the patient has a reduced level of consciousness and alertness. Decerebrate posture is a body pose indicating severe brain damage. In the other chunks, the \emph{DNR} status is again highlighted. Though, the cross-modal network actually does not have a higher likelihood of death for this patient at the time of this note. Hence, it might not have been able to learn about such medical knowledge yet. Interestingly the last note chunk, which strongly indicates that death is near, shows attention on the \emph{namepatterns} and not the patient being placed on comfort measures and that they were extubated (cf. Figure \ref{fig:decompex2chunk32}). Since we did not train end-to-end due to resource constraints, it is possible that the model was unable to capture the expected humanly relevant rationales correctly and represented this observation differently instead.

\end{document}